\documentclass[sigconf]{acmart}
\pdfoutput=1
\renewcommand\footnotetextcopyrightpermission[1]{}
\usepackage{amssymb,amsmath,amsthm}
\usepackage{bm}
\usepackage{microtype}
\usepackage{graphicx}
\usepackage{booktabs}
\usepackage{algorithm}
\usepackage{algpseudocode}

\DeclareMathOperator{\diag}{diag}

\DeclareMathOperator{\tr}{tr}

\newcommand{\dataname}[1]{\textbf{#1}}
\newcommand{\PWF}{P\mkern-0.5mu W\mkern-2mu F}
\newcommand{\PWFPR}{P\mkern-0.5mu W\mkern-2mu F\mkern-1mu P\mkern-1mu R}
\newcommand{\PWFNR}{P\mkern-0.5mu W\mkern-2mu F\mkern-1.5mu N\mkern-1.5mu R}

\theoremstyle{plain}% default

\theoremstyle{definition}

\setcopyright{rightsretained}
\begin{document}
\title[Hybrid Clustering using Joint NMF]{Hybrid Clustering based on Content and Connection Structure using
Joint Nonnegative Matrix Factorization}
\author{Rundong Du}
\affiliation{\institution{Georgia Institute of Technology}
\department{School of Mathematics}
\streetaddress{686 Cherry Street}
\city{Atlanta}
\state{GA}
\postcode{30332-0160}
\country{USA}
}
\email{rdu@gatech.edu}
\author{Barry Drake}
\affiliation{\institution{Georgia Institute of Technology}
\department{Georgia Tech Research Institute}
\department{Information and Communications Lab}
\streetaddress{75 5th Street, N.W., Suite 900}
\city{Atlanta}
\state{GA}
\postcode{30308}
\country{USA}}
\email{barry.drake@gtri.gatech.edu}
\author{Haesun Park}
\affiliation{\institution{Georgia Institute of Technology}
\department{School of Computational Science and Engineering}
\streetaddress{266 Ferst Drive}
\city{Atlanta}
\state{GA}
\postcode{30332-0765}
\country{USA}}
\email{hpark@cc.gatech.edu}

\begin{abstract}
We present a hybrid method for latent information discovery on the data sets
containing both text content and connection structure based on constrained low rank approximation. The new method jointly optimizes the Nonnegative Matrix Factorization (NMF) objective function for text clustering and the Symmetric NMF (SymNMF) objective function for graph clustering.
We propose an effective algorithm for the joint NMF objective function,
based on a block coordinate descent (BCD) framework. The proposed hybrid method discovers content associations via latent connections found using SymNMF. The method can also be applied with a natural conversion of the problem when a hypergraph formulation is used or the content is associated with hypergraph edges.

Experimental results show that by simultaneously
utilizing both content and connection structure,
our hybrid method produces higher quality clustering results
compared to the other NMF clustering methods that uses content alone
(standard NMF) or connection structure alone (SymNMF).
We also present some interesting applications to several types of real world data such as citation recommendations of papers.
%has potential applications to citation recommendations of papers and patents, organization hierarchy detection and team/department detection.
The hybrid method proposed in this paper can also be applied to general data expressed with both feature space vectors and pairwise similarities and can be extended to the case with multiple feature spaces or multiple similarity measures.

\end{abstract}



\maketitle

% !TEX root=hybrid.tex
\section{Introduction}

Constrained low rank approximation (CLRA) such as
Nonnegative matrix factorization (NMF) has played an important role
in data analytics, providing a foundational framework
for formulating key analytics tasks such as text clustering,
graph clustering, and recommendation systems \cite{kuang_fast_2013-1,kuang_symmetric_2012-1,kuang_symnmf:_2014,kuang_nonnegative_2015-1} problems.
In this paper, we propose a joint NMF algorithm which
jointly optimizes the standard NMF for content
clustering and Symmetric NMF (SymNMF) for graph clustering.
Detailed discussions of NMF and SymNMF can be found in \cite{kim_algorithms_2013,kim_fast_2011-1}  and \cite{kuang_symnmf:_2014}, respectively.
The goal
is to cluster data sets that contain both content and connection structure
simultaneously, utilizing both information sources, to obtain higher quality
clustering results. This type of fusion can be done at the data level (early fusion) or at the result level (late fusion). An advantage of NMF and SymNMF is
that both are formulated using one framework of CLRA, and therefore,
we can naturally design a joint objective function to obtain
the objective function level fusion as we illustrate in a later section.

Numerous data sets contain both text content and connection structure.
For example, in a data set of research papers or patents,
papers or patents have text content where the citations or co-author relationships define the connection structure;
in a data set of emails, email messages have text content and the sender-recipient relations define a hypergraph structure where one email may have multiple
recipients.
When we represent the data set as a graph where
the connection structure is represented as edges,
in the former case the text content is associated with graph nodes while in the latter case the text content is associated with hypergraph edges.
For these data sets, clustering based on only text or connection structure
would waste the other source of information.
A hybrid clustering method is designed to utilize both
content and connection structure information, thus taking advantage of the full data context.

Many methodologies exist for data clustering.
However, our framework using CLRA has at least four advantages: (1) simplicity of implementation, widely applicable, and does not assume too much about the data. Although this sometimes means CLRA methods are not as specific or as accurate as more complex and targeted models; (2) CLRA methods can usually provide some valuable insights about the data when there is not enough knowledge about the underlying data model or when one desires only a quick glance at results. In fact, in the area of text and graph clustering, CLRA methods (NMF and SymNMF) have been demonstrated to have superior performance in terms of speed and accuracy \cite{kuang_fast_2013-1,kuang_symmetric_2012-1,kuang_symnmf:_2014}; (3) CLRA methods can be solved by efficient numerical algorithms and have sophisticated numerical linear algebra and optimization algorithms/libraries such as BLAS and LAPACK as a foundation; (4) the two CLRA methods (NMF for text and SymNMF for graph clustering) have the same underlying matrix factorization framework, and, therefore, have consistent interpretations, which makes it more straightforward to combine the two.

The use of joint matrix factorization for clustering can also be seen in  \cite{liu_multi-view_2013,tang_integrating_2012,zhu_combining_2007,jin_combined_2015}, all of which consider clustering using information from different sources. \cite{zhu_combining_2007} and \cite{tang_integrating_2012} are also methods for hybrid clustering of connection structure and content data. However, \cite{zhu_combining_2007} used a different objective function, which did not have nonnegative constraints. \cite{tang_integrating_2012}  did not consider symmetric factorization of the adjacency matrix and used different constraints. Also,
\cite{liu_multi-view_2013} did not consider graph data and therefore symmetric factorization is not incorporated. \cite{jin_combined_2015} used a similar objective function as ours, but their method was used only for graph clustering.

Other methods for hybrid clustering include generative models \cite{cohn_missing_2001,erosheva_mixed-membership_2004,chang_hierarchical_2010,Gruber08,liu_topic-link_2009,nallapati_joint_2008}, topic modeling with network regularization \cite{mei_topic_2008,sun_relation_2012}, augmenting the graph with content information \cite{ruan_efficient_2013}, an entropy based method \cite{cruz_entropy_2011}, cluster ensembles \cite{strehl_cluster_2003}, and cluster selection \cite{elhadi_structure_2013}.

In this paper we discuss data with associated text content and connection structure. In addition to text content, other types of information may also be associated with connection structure.
The information falls in two categories:
text content and images,
and attributes that appear in structured data,
as in a database, such as a persons age and gender, etc.
Our hybrid clustering method can naturally extend to other
content information as long as the raw data can be encoded as nonnegative vectors.
However, our CLRA framework may not be
suitable for attribute information, which can be encoded in
very low dimensional vectors.
Therefore, in our study, we do not include approaches
designed only for attributes.

This paper is organized as follows: We start with the basic situation where the text content are associated with connection structure, i.e.,graph nodes, and extend the idea to the case where a hypergraph is the correct connection representation and the text content is associated with hypergraph edges (Section~\ref{sec:methods}). We have conducted extensive experiments using patent citation data to show the effectiveness of our method (Section~\ref{sec:experiments}). In addition to demonstrating improvements of clustering quality, we list several potential applications of our hybrid clustering approach, including the application of our hypergraph extension on an Enron email data set (Section~\ref{sec:application}). Discussions and conclusions can be found in Section~\ref{sec:conclusion}.

% !TEX root=hybrid.tex
\section{Hybrid Clustering via Joint NMF}
\label{sec:methods}

We have designed fast, scalable algorithms
for some variants of NMF for key data analytics problems
\cite{kim_fast_2011-1, kuang_fast_2013-1, utopian2013}.
Currently one of the fastest algorithms for hierarchical
and flat (non-hierarchical) topic modeling and clustering
that also produce consistently high quality solutions are
HierNMF2 and FlatNMF2, which are available in our open source software
package in C++ called SmallK  (\url{http://smallk.github.io/}).

First we assume that the text content are associated with the graph nodes (e.g.\ paper/patents with citations).
We assume that a data set's text information
is represented in a nonnegative matrix
$X\in\mathbb{R}_{+}^{m\times n}$ and the graph structure
is represented in a nonnegative symmetric matrix
$S \in \mathbb{R}_+^{n \times n}$,
where $m$ is the number of features,
the columns of $X$ represent the $n$ data items, the $(i,j)$th element of $S$ represents
a relationship such as similarity between the $i$th and $j$th data items,
and $\mathbb{R}_+$ denotes the real nonnegative numbers.
Then the NMF formulation for text clustering/topic modeling \cite{kuang_nonnegative_2015-1} is
\begin{equation}
  \label{eq:nmf}
  \min_{W\ge 0, H\ge 0}\|X-WH\|_F
\end{equation}
 and the SymNMF formulation for graph clustering \cite{kuang_symmetric_2012-1,kuang_symnmf:_2014} is
\begin{equation}
  \label{eq:symnmf}
 \min_{H\ge0} \|S-H^TH\|_F
\end{equation}
where $W\in\mathbb{R}_{+}^{m\times k}$ and $H\in\mathbb{R}_{+}^{k\times n}$,
and a given integer $k$, which is typically much smaller than $m$ or $n$, represents the reduced dimension, i.e., number of clusters \cite{kim_algorithms_2013}.
In \eqref{eq:nmf}, each column of $W$, subject to some scaling,
is regarded as the representative of each cluster or a topic in the document collection. The matrix $H$ can be seen as a low rank (rank $k$) representation of the data points since each data item in $X$ can be explained
by an additive linear combination of the representative columns in $W$, i.e., the columns of $H$ are approximative coordinates of data items in $X$ with columns of $W$ as basis vectors.
 Similarly, in \eqref{eq:symnmf}, $H$ is a low rank representation of the nodes in the graph. Such a low rank approximation also gives us $k$ clusters,
since $H_{i,j}$ can be seen as a measurement of
strength that the $j$th data item belongs to the $i$th cluster. Therefore, each column of $H$ gives the soft clustering assignment information.
By taking the row index with the maximum value in each column
vector of $H$ as the cluster index of each data item, one can also perform hard
clustering  \cite{kim_algorithms_2013,kim_fast_2011-1}.

The hybrid clustering method we propose finds a low rank representation that simultaneously represents the text content and the graph structure of the data items by jointly optimizing the NMF and SymNMF objective functions:
\begin{equation}
  \label{eq:jointnmf-general-0}
  \min_{W\geq 0, H\geq 0} \alpha_1|| X-WH ||_F^2 + \alpha_2 || S-H^T H ||_F^2.
\end{equation}
where $\alpha_1\ge 0$ and $\alpha_2 \ge 0$ are the weighting parameters. By adjusting the parameters $\alpha_i$, we can emphasize
one over the other. In the extreme case, some $\alpha_i$
can be set to zero: e.g. when $\alpha_2=0$ in the above,
we are only concerned with the content,
when $\alpha_1=0$, we only pay attention to the structural
information and ignore the content.
Excluding these special cases, we  can assume $\alpha_1=1$ without loss of generality and Eqn.~\eqref{eq:jointnmf-general-0} becomes
\begin{equation}
  \label{eq:jointnmf}
  \min_{W\geq 0, H\geq 0} || X-WH ||_F^2 + \alpha || S-H^T H ||_F^2.
\end{equation}
with $\alpha\ge 0$ as the weighting parameter.

Now we extend our method to hypergraphs where the text content is associated with hypergraph nodes.  Once this is done, it would be natural to extend our method further to the cases where text is associated with graph
or hypergraph edges due to the duality that exists between edges and nodes of a  hypergraph and the fact that a graph can be treated as a special case of a hypergraph.

A hypergraph $\mathcal{H}$ is a pair $\mathcal{H} = (\mathcal{V},\mathcal{E})$, where $\mathcal{V}=\{v_1,\ldots,v_m\}$ is the set of vertices and $\mathcal{E}=\{e_1,\ldots,e_n:\,e_i\in\mathcal{V}\}$ is the set of hyperedges. Unlike a graph edge, a hypergraph edge $e_i$ may connect more than two vertices in the graph. Such a hypergraph $\mathcal{H}$ can be represented by an incidence matrix $M=(m_{ij})\in\mathbb{R}^{m\times n}$,
where
\begin{equation*}
  m_{ij} =
  \begin{cases}
    1,& v_i\in e_j;\\
    0,& \text{otherwise.}
  \end{cases}
\end{equation*}
The dual hypergraph $\mathcal{H}^*$ is the hypergraph corresponding to the incidence matrix $M^T$.

Assume there's a $k$-way partition of the vertices $(\mathcal{V}_1,\ldots,\mathcal{V}_k)$ where
$\mathcal{V}_1 \cup \cdots \cup \mathcal{V}_k=\mathcal{V}$ and
$\mathcal{V}_i \cap \mathcal{V}_j=\emptyset$ for all $1\le i\neq j\le k$. Define the matrix $H=(h_{ij})\in\mathbb{R}^{k\times n}$ as
\begin{equation}
h_{ij} = \frac{[v_j\in \mathcal{V}_i]}{\sqrt{d_v(j)}\left(\displaystyle\sum_{v_l\in\mathcal{V}_i}\frac{1}{d_v(l)}\right)^{1/2}}
\label{eq:H-discrete}
\end{equation}
which is a normalized partition indicator matrix where
\begin{equation*}
  [v_j\in\mathcal{V}_i] =
  \begin{cases}
    1, & v_j\in\mathcal{V}_i;\\
    0, & \text{otherwise.}
  \end{cases}
\end{equation*}
and $d_v(l) = \sum_{j=1}^n m_{lj}$ is the degree of vertex $v_l$.
It is shown in \cite{zhou_learning_2007} that the following optimization problem
\begin{equation}
  \max_H \tr HSH^T
  \label{eq:maxtr}
\end{equation}
is equivalent to minimizing the hypergraph normalized cut as defined in \cite{zhou_learning_2007},
where
\begin{equation}
  S=D_v^{-1/2}MD_e^{-1}M^TD_v^{-1/2}
  \label{eq:S}
\end{equation}
 is symmetric, $D_v=\diag(d_v(1),\ldots,d_v(m))$, \\ $D_e=\diag(d_e(1),\ldots,d_e(n))$, and $d_e(l) = \sum_{i=1}^m m_{il}$ is the degree of edge $e_l$.
Following the same argument as in \cite{kuang_symmetric_2012-1},
it can be shown that \eqref{eq:maxtr} is equivalent to  $\min_{H} \|S-H^TH\|_F^2$ and by relaxing constraint \eqref{eq:H-discrete} to $H\ge 0$,
we obtain the objective function of SymNMF. Therefore, in the case of a hypergraph, we can use the matrix $S$ defined in Eqn.~\eqref{eq:S} as the similarity matrix in Eqn.~\eqref{eq:jointnmf}.

There are many ways to find a solution for the
objective function \eqref{eq:jointnmf}.
We propose reformulating it in the following form with a penalty term
\begin{equation}
  \label{eq:jointnmf-reg}
  \min_{W,H,\tilde{H}\ge 0} || X-WH ||_F^2 + \alpha || S-\tilde{H}^T H ||_F^2 + \beta\|\tilde{H}-H\|_F^2.
\end{equation}
where $\tilde{H}\in\mathbb{R}_+^{k\times n}$ and $\beta\ge 0$ is the regularization parameter. This reformulation is motivated from our earlier work
to generate an algorithm that is based on the block coordinate descent (BCD) scheme
so that each sub-problem in the BCD is a nonnegativity constrained least
squares (NLS) problem for which we have developed a highly efficient algorithm
and optimized open-source software \cite{SmallK}.
Then Eqn.~\eqref{eq:jointnmf-reg} can be solved using a 3-block coordinate descent (BCD) scheme, i.e. minimize the objective function with respect to $W$, $\tilde{H}$ and $H$ in turn. Specifically, we solve the following three subproblems in turn:
\begin{align}
  &\min_{W\ge 0} \|H^TW^T-X^T\|_F^2\\
  &\min_{\tilde{H}\ge 0} \left\|
  \begin{bmatrix}
    \sqrt{\alpha} H^T\\
    \sqrt{\beta} I_k
  \end{bmatrix}\tilde{H}
  -
  \begin{bmatrix}
    \sqrt{\alpha} S\\
    \sqrt{\beta} H
  \end{bmatrix}
  \right\|_F^2\\
  &\min_{H\ge 0}\left\|
  \begin{bmatrix}
    W\\\sqrt{\alpha}\tilde{H}^T\\\sqrt{\beta} I_k
  \end{bmatrix} H -
  \begin{bmatrix}
    X\\\sqrt{\alpha}S\\\sqrt{\beta} \tilde{H}
  \end{bmatrix}
  \right\|_F^2
\end{align}
where each subproblem is simply a nonnegative least squares problem (NLS),
which is convex. Thus, an active-set-based algorithm can
 find the optimal solution in a finite number of operations and ensures that the solution is in the feasible region. Thus,
 avoiding the case of nearly linear dependent vectors, which has profound implications for real-world applications
 such as chemical detection where false negatives and false positives can increase dramatically in the presence of
 rank deficiency \cite{DKMP10}. The three block BCD algorithm converges to a stationary point according to Bertsekas' theorem \cite{Bertsekas1999}.
The identity submatrices $I_k$ in the above equations make
the problem better conditioned than the subproblems in the standard NMF
that uses two block BCD alternating updating $W$ and $H$.
We solve each NLS problem using the block principal pivoting (BPP) algorithm \cite{kim_fast_2011-1}. Theoretically, to force $H$ to be identical to $\tilde {H}$, the value of the parameter $\beta$ has to be infinity.
This problem has been studied extensively and we use
a scheme similar to what was proposed in \cite{admm2012}. It should be pointed out that also in \cite{kim_fast_2011-1} it is shown that algorithms based on the  BCD framework have guaranteed convergence to a stationary point, whereas, popular and easy to implement algorithms such as Multiplicative Updating (MU) may not converge. In addition, extensive experiments show that the BPP method is faster and more accurate than MU.

% !TEX root = hybrid.tex
% \subsection{Data Overview}
% The data we used for experiments include US patent claim and citation data from PatentsView\footnote{\url{http://www.patentsview.org}}, paper title/abstract and citation data \dataname{cit-HepTh} from SNAP\cite{snapnets}, and a subset of Enron email data from UC Berkeley \footnote{\url{http://bailando.sims.berkeley.edu/enron_email.html}}.

%\subsection{Experiment Design}

\section{Clustering US Patent Data}
\label{sec:experiments}
All experiments were performed on a server with two Intel(R) Xeon(R) CPU E5-2680 v3 CPUs and 377GB memory.
We use US patent claim and citation data from PatentsView\footnote{\url{http://www.patentsview.org}}. Some advantages of using US patents as a data source are: (1) the openness, centralized management and availability of relatively structured data format makes the patent data easier to obtain and process; (2) the abundance of the patent database ensures enough samples that can be studied; (3) patents were carefully assigned with classification labels, and such labels were examined by patent examiners; therefore the classification information can be used as a relatively reliable ground truth.

\begin{figure}[htp]
  \centering
  \includegraphics{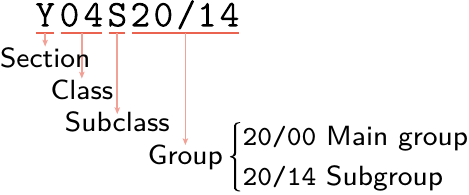}
  \caption{An example classification label in the CPC scheme}
  \label{fig:cpc}
\end{figure}

We used the Cooperative Patent Classification (CPC) system, where each classification label has the scheme as illustrated in Figure~\ref{fig:cpc}.
We select 13 CPC classes (A22, A42, B06, B09, B68, C06, C13, C14, C40, D02, D10, F22, Y04) and use patents under each class to construct 13 different data sets. For each data set, we first construct the term-document matrix representing the patent claims and the graph adjacency matrix representing the patent citation relations. Our algorithm requires a symmetric adjacency matrix and therefore we treat the citation graph as undirected by ignoring the directions. We then clean the data by removing terms that appear very infrequently and documents that are too short or duplicated, and extracting the largest connected components of the graph. Finally, we apply tf-idf to the term-document matrix, normalize its columns to have unit 2-norm, obtaining the matrix $X$, and let $S$ be $D^{-1/2}AD^{-1/2}$, where $A\in\mathbb{R}^{n\times n}$ is the adjacency matrix, $D=\diag(d_1,\ldots,d_n)$ and $d_i=\sum_{j=1}^nA_{ij}$ is the degree of vertex $i$. We use CPC groups as ground truth clusters. Some statistics about these data sets (after cleaning) are listed in Table~\ref{tab:patent-subset}.
\begin{table}
  \caption{Some statistics of US patent data sets.}
  \label{tab:patent-subset}
  \begin{tabular}{rrrr}
    \toprule
    Class & \#Patents & \#Citations & \#Groups\\
    \midrule
    A22&4976&28746&230\\
    A42&4213&29285&134\\
    B06&2938&11549&82\\
    B09&3522&17302&38\\
    B68&790&2433&93\\
    C06&3347&17562&141\\
    C13&1010&3717&87\\
    C14&583&1125&69\\
    C40&3748&28854&41\\
    D02&3170&11216&158\\
    D10&2548&8486&154\\
    F22&3040&7977&359\\
    Y04&3242&21518&76\\
    \bottomrule
  \end{tabular}
\end{table}

We now define the measures for the evaluation of the clustering results. Assume we computed $k$ clusters $B_1,\ldots,B_k$ and the ground truth has $k'$ clusters
$G_1,\ldots,G_{k'}$. We compute the confusion matrix $C=(c_{ij})_{k\times k'}$, where $c_{ij}=|A_i\cap B_j|$. Then we define the \emph{average $F_1$ score} \cite{yang_overlapping_2013} as
\begin{equation*}
  F_1=\frac{1}{2}\left(
    \frac{1}{k}\sum_{i=1}^k\max_jF_1(A_i,B_j)
    +
    \frac{1}{k'}\sum_{j=1}^{k'}\max_iF_1(B_j,A_i)
  \right)
\end{equation*}
where
\begin{equation*}
  F_1(A_i,B_j)=F_1(B_j,A_i)=\frac{2c_{ij}}{|A_i|+|B_j|}
\end{equation*}
This score measures how well an algorithm can recover the ground truth clusters. We also define another type of $F_1$ score called \emph{pairwise $F_1$ score}, as seen in \cite{manning_introduction_2008,yang_combining_2009}, which measures how well an algorithm can predict the connections among data items. Assume there are $n$ data items in total. For each of the $n(n-1)/2$ pairs of data items, we say the two items are \emph{c-connected} if they belong to the same cluster, otherwise we call them \emph{c-disconnected} (a prefix c is added to distinguish from connectivity in graph theory). Clustering results can also be treated as a prediction of c-connectivity of each pair of data items.
A prediction regarding one pair of data items can have
four cases of true positive (TP), true negative (TN), false positive (FP) or false negative (FN) according to the rules listed in Table~\ref{tab:tfpn}.
\begin{table}
  \caption{Type of predictions}
  \label{tab:tfpn}
  \centering
  \begin{tabular}{lll}
    \toprule
    In prediction & In ground truth & Type\\
    \midrule
    c-connected & c-connected & TP\\
    c-disconnected & c-disconnected & TN \\
    c-connected & c-disconnected & FP\\
    c-disconnected & c-connected & FN \\
    \bottomrule
  \end{tabular}
\end{table}
We then define pairwise $F_1$ score ($\PWF_1$) as
\begin{equation*}
  \PWF_1 = \frac{2\#TP}{2\#TP+\#FN+\#FP}
\end{equation*}
To study the type of errors each algorithm makes, we also define pairwise false positive rate ($\PWFPR$) and pairwise false negative rate ($\PWFNR$) as
\begin{align*}
  &\PWFPR = \frac{\#FP}{\#FP+\#TN}\\
  &\PWFNR = \frac{\#FN}{\#FN+\#TP}
\end{align*}
Note that in the case of average F1 score, there are no real false positives or false negatives because we are actually matching detected clusters with ground truth clusters and put them in symmetric positions. Therefore, we don't discuss these two rates there. To evaluate pairwise scores, we also utilize connection information from external patent classifications. For example, patents in the class Y04 may also have classification labels in B06 and B09. Those external labels do not form a complete cluster, therefore we exclude them when evaluating the cluster quality. But they contain valuable connection information among different patents, therefore we include them in the pairwise scores.

\begin{figure*}[!tp]
  \centering
  \begin{minipage}{\columnwidth}
  \includegraphics[width=\linewidth]{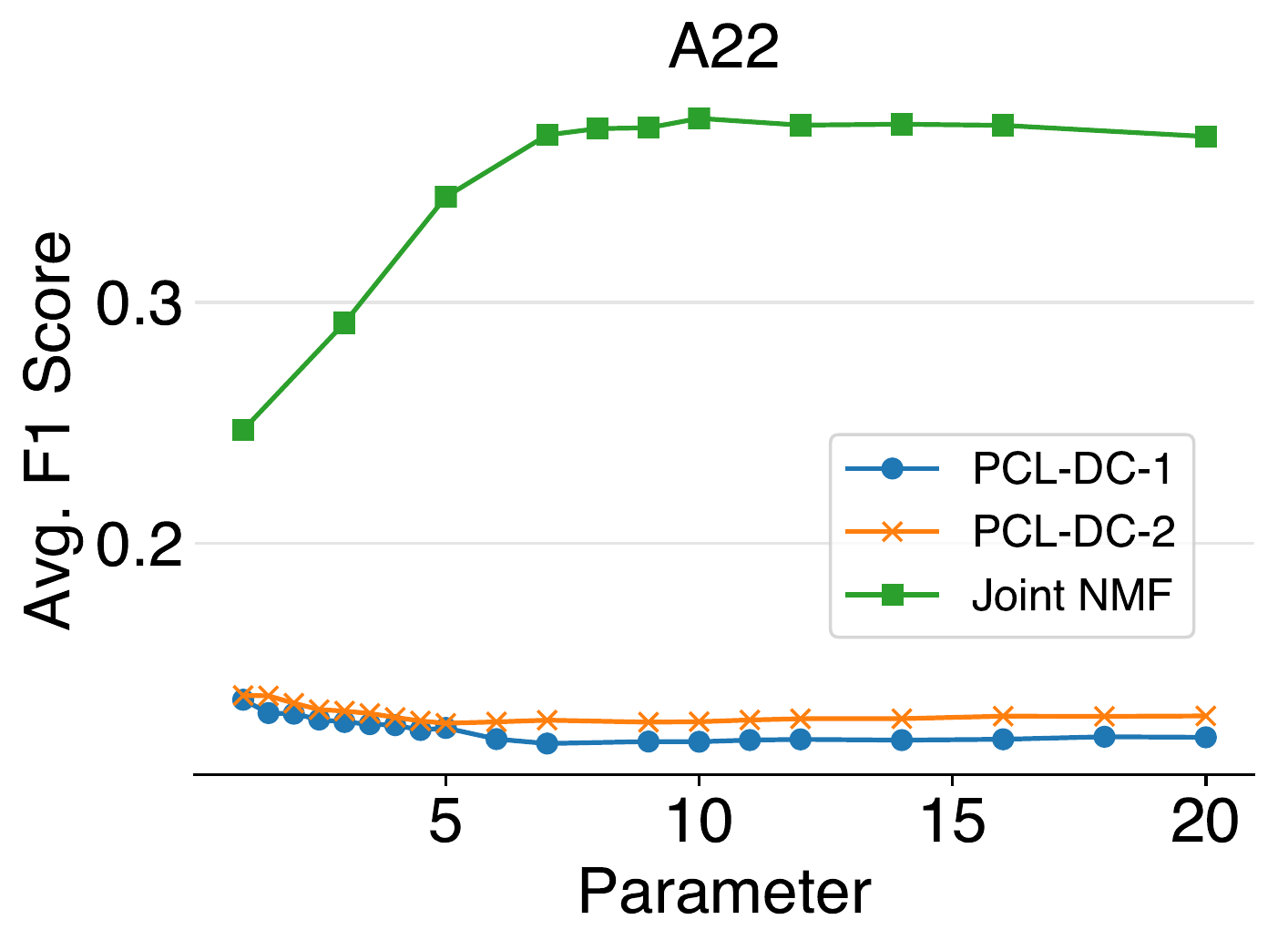}
  \end{minipage}
  \begin{minipage}{\columnwidth}
  \includegraphics[width=\linewidth]{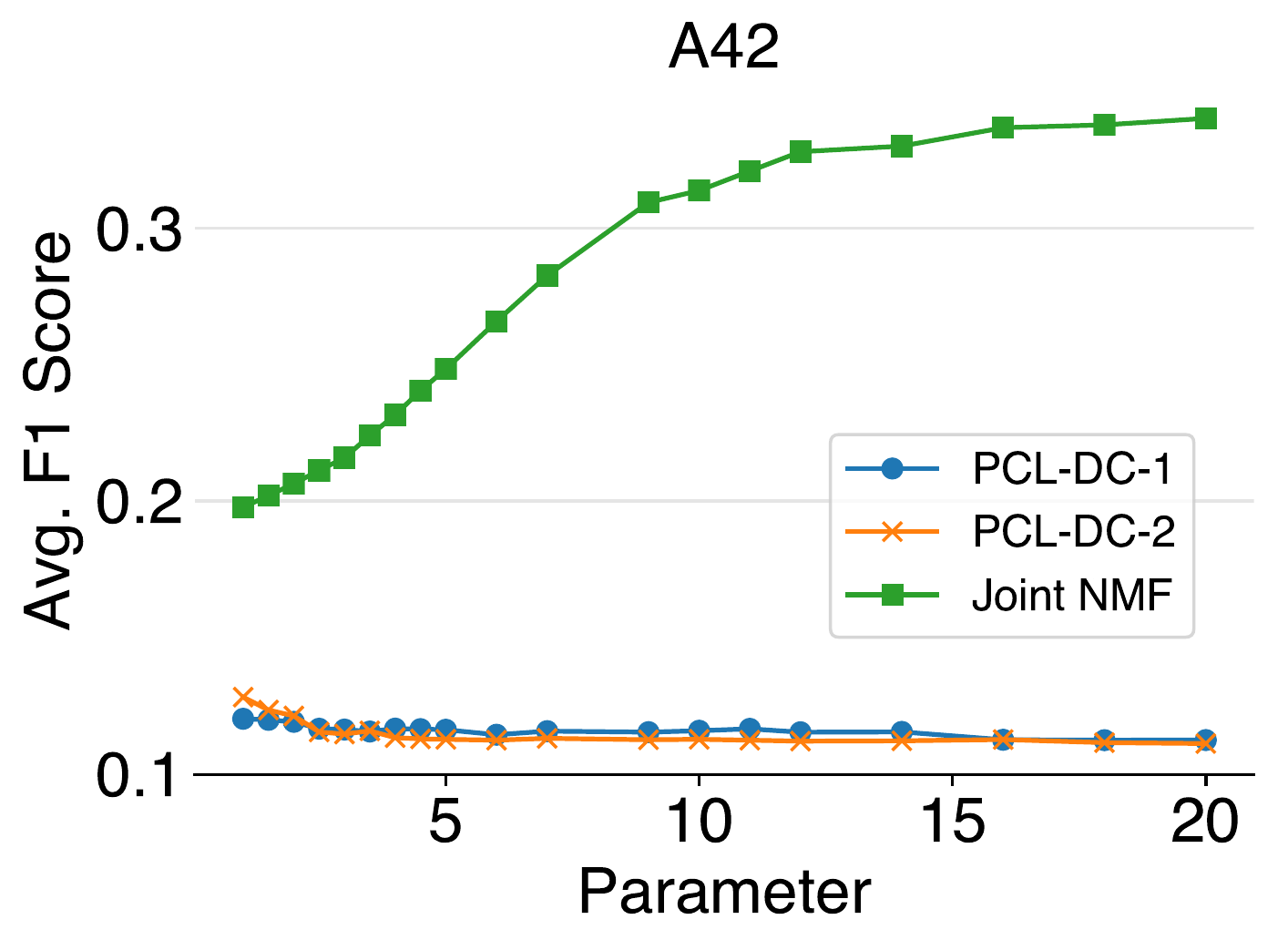}
  \end{minipage}
  \begin{minipage}{\columnwidth}
  \includegraphics[width=\linewidth]{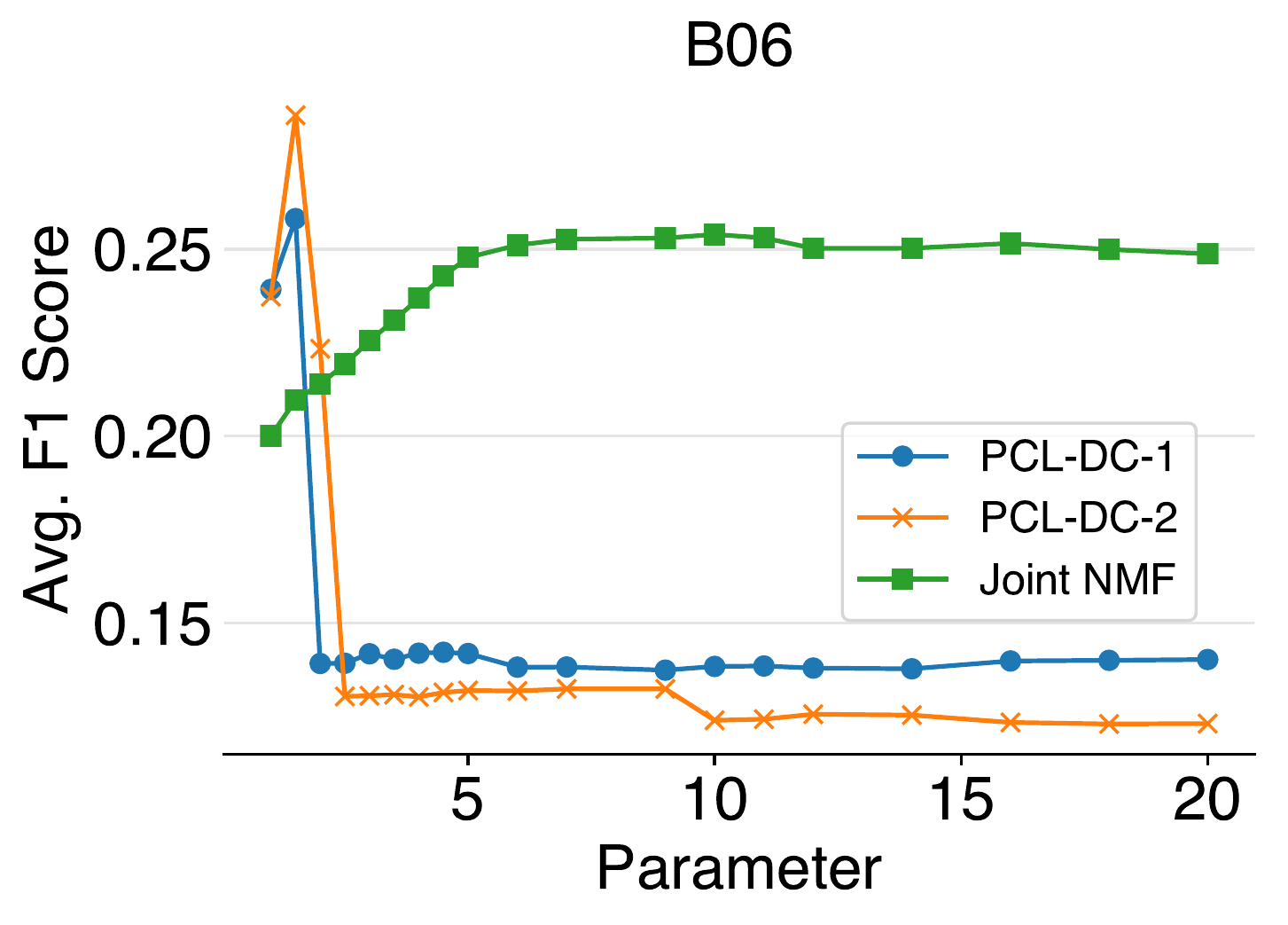}
  \end{minipage}
  \begin{minipage}{\columnwidth}
  \includegraphics[width=\linewidth]{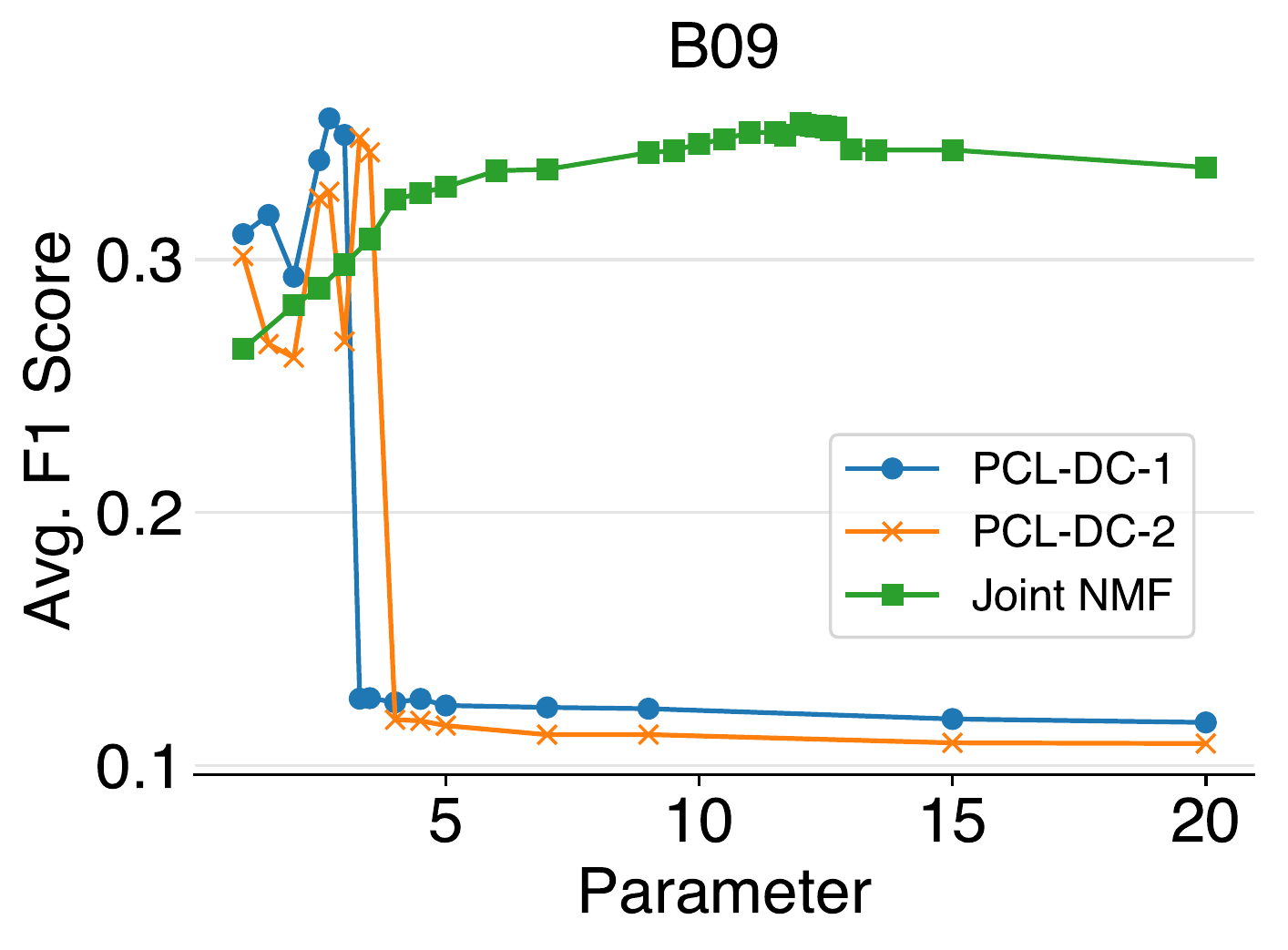}
  \end{minipage}
  \caption{Parameter sensitivity of PCL-DC and Joint NMF.}
  \label{fig:params}
\end{figure*}

We compare our algorithm with NMF and SymNMF, which have leading performance in text clustering and graph clustering, respectively. For hybrid clustering, we choose PCL-DC \cite{yang_combining_2009} to compare with based on popularity and source code availability.
Both joint NMF and PCL-DC have parameters to set. For joint NMF, we let the default parameter to be $\alpha=\|X\|_F^2/\|S\|_F^2$, standing for a half-half balance between graph clustering and text clustering, and we set $\beta=\alpha\|S\|_{max}$, where $\|S\|_{max}$ is the maximum absolute value of elements in $S$. The authors of PCL-DC do not provide a way to specify its regularization parameter $\lambda$. Therefore, we need to first study how parameter change will affect the algorithm performance. We found that for $\lambda<1$, PCL-DC sometimes becomes extremely slow, such that it may take weeks to run it over all the data sets (estimated based on sampling run). Therefore, we let $\lambda$ vary within $[1,20]$. In Figure~\ref{fig:params}, we show how the average F1 score changes when $\lambda$ varies in that range for the first four data sets listed in Table~\ref{tab:patent-subset}. The code of PCL-DC\footnote{\url{https://homepage.cs.uiowa.edu/~tyng/codes/community_detection.zip}} provides two models (popularity link model and productivity link model), and we call them PCL-DC-1 and PCL-DC-2, respectively. We also show the performance change of joint NMF when its parameter $\alpha$ varies in the same range.
We can observe that the PCL-DC is either worse than joint NMF or very sensitive to the parameters, and it seems that when $\lambda$ exceed a certain threshold (depending on the data), there will be a big drop in clustering quality. Therefore, to have a tolerable running time while having a fair clustering quality, we choose $\lambda=1$ in the comparison experiments. The results of the comparison are listed in Table~\ref{tab:f1} to Table~\ref{tab:pwfnr}, where each value is the average over 10 runs.
\begin{table}
  \caption{Comparison of average F1 scores}
  \label{tab:f1}
  \setlength{\tabcolsep}{4pt}
  \centering
  \begin{tabular}{cccccc}
\toprule
Class &  Joint NMF &    NMF &  SymNMF &  PCL-DC-1 &  PCL-DC-2 \\
\midrule
A22  &     \textbf{0.3730} & 0.2293 &  0.3457 &    0.1351 &    0.1369 \\
A42  &     \textbf{0.3215} & 0.1779 &  0.3199 &    0.1201 &    0.1280 \\
B06  &     \textbf{0.2502} & 0.1905 &  0.2307 &    0.2393 &    0.2373 \\
B09  &     \textbf{0.3336} & 0.2449 &  0.2690 &    0.3101 &    0.3014 \\
B68  &     0.3806 & 0.3059 &  0.3762 &    \textbf{0.4034} &    0.3671 \\
C06  &     \textbf{0.2257} & 0.1830 &  0.2004 &    0.1156 &    0.1158 \\
C13  &     \textbf{0.2990} & 0.2664 &  0.2953 &    0.2616 &    0.2224 \\
C14  &     \textbf{0.3584} & 0.3191 &  0.3578 &    0.2692 &    0.2659 \\
C40  &     0.1939 & 0.1709 &  0.1673 &    0.1951 &    \textbf{0.1981} \\
D02  &     \textbf{0.2990} & 0.2131 &  0.2683 &    0.1756 &    0.2268 \\
D10  &     \textbf{0.3046} & 0.2452 &  0.2783 &    0.1612 &    0.2999 \\
F22  &     \textbf{0.3006} & 0.2211 &  0.2926 &    0.1533 &    0.1388 \\
Y04  &     0.2489 & 0.2069 &  0.2018 &    \textbf{0.2599} &    0.2596 \\
\bottomrule
\end{tabular}
\end{table}
\begin{table}
  \caption{Comparison of pairwise F1 scores}
  \label{tab:pwf1}
  \setlength{\tabcolsep}{4pt}
  \begin{tabular}{cccccc}
\toprule
Class &  Joint NMF &    NMF &  SymNMF &  PCL-DC-1 &  PCL-DC-2 \\
\midrule
A22  &     \textbf{0.2814} & 0.1310 &  0.2493 &    0.1091 &    0.1108 \\
A42  &     \textbf{0.2697} & 0.0947 &  0.2434 &    0.1104 &    0.1021 \\
B06  &     0.1777 & 0.1115 &  0.3703 &    \textbf{0.4083} &    0.3652 \\
B09  &     0.2212 & 0.1439 &  \textbf{0.4080} &    0.3173 &    0.3254 \\
B68  &     0.2821 & 0.1633 &  0.3281 &    \textbf{0.5252} &    0.4322 \\
C06  &     0.1324 & 0.0726 &  \textbf{0.2092} &    0.1343 &    0.1924 \\
C13  &     0.1457 & 0.0979 &  0.2001 &    0.3570 &    \textbf{0.4394} \\
C14  &     0.1558 & 0.1290 &  0.1559 &    0.2537 &    \textbf{0.2569} \\
C40  &     0.1274 & 0.0909 &  \textbf{0.4431} &    0.1832 &    0.2149 \\
D02  &     0.1200 & 0.0725 &  0.1430 &    0.2407 &    \textbf{0.2549} \\
D10  &     0.0905 & 0.0560 &  0.1046 &    0.1519 &    \textbf{0.2686} \\
F22  &     \textbf{0.1850} & 0.0746 &  0.1720 &    0.1342 &    0.1044 \\
Y04  &     0.1767 & 0.0851 &  0.3681 &    0.4232 &    \textbf{0.4510} \\
\bottomrule
\end{tabular}
\end{table}
\begin{table}
  \caption{Comparison of pairwise false positive rates}
  \label{tab:pwfpr}
  \setlength{\tabcolsep}{4pt}
  \begin{tabular}{cccccc}
    \toprule
Class &  Joint NMF &    NMF &  SymNMF &  PCL-DC-1 &  PCL-DC-2 \\
\midrule
A22  &     0.0052 & \textbf{0.0044} &  0.0060 &    0.0569 &    0.0342 \\
A42  &     0.0109 & \textbf{0.0078} &  0.0105 &    0.0525 &    0.0422 \\
B06  &     0.0114 & \textbf{0.0106} &  0.0601 &    0.0967 &    0.0887 \\
B09  &     \textbf{0.0187} & 0.0225 &  0.1924 &    0.0371 &    0.0584 \\
B68  &     \textbf{0.0043} & 0.0071 &  0.0058 &    0.0592 &    0.0995 \\
C06  &     0.0056 & \textbf{0.0054} &  0.0099 &    0.0300 &    0.0351 \\
C13  &     \textbf{0.0068} & 0.0082 &  0.0104 &    0.0721 &    0.1068 \\
C14  &     \textbf{0.0084} & 0.0125 &  0.0100 &    0.1244 &    0.1658 \\
C40  &     \textbf{0.0147} & 0.0149 &  0.0959 &    0.0311 &    0.0286 \\
D02  &     \textbf{0.0053} & \textbf{0.0053} &  0.0098 &    0.0863 &    0.0954 \\
D10  &     \textbf{0.0032} & 0.0043 &  0.0046 &    0.0528 &    0.0530 \\
F22  &     0.0033 & 0.0033 &  \textbf{0.0026} &    0.0296 &    0.0295 \\
Y04  &     0.0144 & \textbf{0.0109} &  0.0580 &    0.0614 &    0.0732 \\
\bottomrule
\end{tabular}
\end{table}
\begin{table}
  \caption{Comparison of pairwise false negative rates}
  \label{tab:pwfnr}
  \setlength{\tabcolsep}{4pt}
  \begin{tabular}{cccccc}
    \toprule
    Class &  Joint NMF &    NMF &  SymNMF &  PCL-DC-1 &  PCL-DC-2 \\
    \midrule
    A22  &     0.7992 & 0.9165 &  0.8203 &    \textbf{0.7986} &    0.8536 \\
    A42  &     \textbf{0.7994} & 0.9401 &  0.8230 &    0.8606 &    0.8863 \\
    B06  &     0.8895 & 0.9337 &  0.6138 &    \textbf{0.4547} &    0.5459 \\
    B09  &     0.8641 & 0.9138 &  \textbf{0.4991} &    0.7768 &    0.7493 \\
    B68  &     0.8252 & 0.9017 &  0.7866 &    0.3292 &    \textbf{0.3151} \\
    C06  &     0.9249 & 0.9602 &  0.8708 &    0.9050 &    \textbf{0.8537} \\
    C13  &     0.9177 & 0.9456 &  0.8808 &    0.6756 &    \textbf{0.5128} \\
    C14  &     0.9092 & 0.9234 &  0.9080 &    0.6939 &    \textbf{0.6351} \\
    C40  &     0.9311 & 0.9518 &  \textbf{0.6880} &    0.8964 &    0.8765 \\
    D02  &     0.9317 & 0.9597 &  0.9129 &    0.7066 &    \textbf{0.6691} \\
    D10  &     0.9518 & 0.9705 &  0.9434 &    0.8940 &    \textbf{0.7998} \\
    F22  &     0.8842 & 0.9561 &  0.8958 &    \textbf{0.8408} &    0.8784 \\
    Y04  &     0.8956 & 0.9530 &  0.7055 &    0.6448 &    \textbf{0.5966} \\
    \bottomrule
\end{tabular}
\end{table}
We can observe that (1) Joint NMF in general has the best average F1 scores, and its average F1 score is better than that of NMF or SymNMF alone, consistently; (2) Each algorithm (except NMF) achieve the best pairwise F1 score several times; (3) Joint NMF and NMF have very low false positive rates, compared to other algorithms; (4) PCL-DC-1 and PCL-DC-2 have lower false negative rates than other algorithms. Note that all these algorithms have relatively high false negative rates. This is because the ground truth information used for pairwise scores contains external classification information and is thus highly overlapping. However, all these algorithms are non-overlapping clustering algorithms, which means many connections between data items cannot be recovered.
In conclusion, the joint NMF produced better quality solutions for clustering;
for prediction of pairwise connection, joint NMF and PCL-DC performed well.
The joint NMF method has other advantages: its parameter has explicit meanings (weight between text and graph), the clustering quality is not very sensitive against the parameter, and its default parameter works very well.

% !TEX root = hybrid.tex
\section{Other Applications}
\label{sec:application}
Besides clustering, joint NMF has other potential applications such as citation recommendations of papers/patents and activity/leader detection in an organization.

\subsection{Citation recommendation}

When applied to papers/patents with citations or web pages with hyperlinks, the formulation \eqref{eq:jointnmf} can also be understood as finding a basis $W$ for the text space, such that under this basis, the representation (coordinates) of the documents can also reflect
their linkage information. Therefore, when we express a new vector $\bm{x}$ in the text space using the basis $W$, i.e.\ finding a vector $\bm{h}$ that solves the following optimization problem
\begin{equation}
  \label{eq:cit}
  \min_{\bm{h}\ge 0}\|\bm{x}-W\bm{h}\|_2
\end{equation}
We can use closeness of $\bm{h}$ to the column vectors in $H$ to decide how likely the new document represented by $\bm{h}$ should cite some of the documents in $H$. For example, one can recommend a new document to cite the $i$-th original document if the $i$-th entry of $H^T\bm{h}$ is larger than certain threshold. Another way is to set the threshold on the cosine similarity between $\bm{h}$ and column vectors in $H$. We will see that each method has its advantages.

For this task, we used the paper title/abstract and citation data \dataname{cit-HepTh} from SNAP\cite{snapnets}, which contains  27,770 papers from January 1993 to April 2003 in the hep-th (high energy physics - theory) section of arXiv. Note that this is a different task from clustering and therefore the data preprocessing procedure is a little different: we use the raw adjacency matrix for $S$ (i.e. $S=A$). The normalized version $D^{-1/2}AD^{-1/2}$ is related to minimizing normalized cut\cite{kuang_symmetric_2012-1} and therefore good for clustering. Here the raw adjacency matrix is a better indicator of citations, which is used as an input that the algorithm learns from, instead of a basis for clustering.

To evaluate our method, we separate the data into training and test sets by treating papers published earlier than 2003 as training set and papers published in 2003 as test set. We learn a matrix $W$ from the documents and citation relation in the training set, make prediction of citations for documents in the test set and compare the prediction with the actual citations.

To verify that the $W$ computed by our algorithm indeed reflects the network structure better, we also designed several baseline methods. A naive method is to predict citations based on number of words shared by two documents. One method based on NMF is to learn the matrix $W$ used in \eqref{eq:cit} only by NMF, i.e. $\min_{W\ge 0, H\ge 0}\|X_{train}-WH\|_F$. Another method based on NMF is to directly learn the $\bm{h}$ vector in \eqref{eq:cit} by $\min_{W,H,\bm{h}\ge 0}\|[X_{train},\bm{x}]-W[H,\bm{h}]\|_F$.
For the two NMF-based methods, the rest of the steps for making predictions are the same as joint NMF, once the matrix $W$ or the vector $\bm{h}$ is obtained. In this subsection, we denote these two NMF based methods as NMF-1 and NMF-2, respectively.

In both prediction methods (compute $H^T\bm{h}$, the inner product , or compute cosine similarity scores), a threshold is needed. Instead of evaluating these algorithms with a fixed threshold, we show the receiver operating characteristic (ROC) curve, which plots the true positive rate against the false positive rate at various threshold values. In general, the closer the curve is to the upper left corner of the graph, the better the algorithm is.

\begin{figure*}[!tp]
  \centering
  \begin{minipage}{\columnwidth}
  \includegraphics[width=\linewidth]{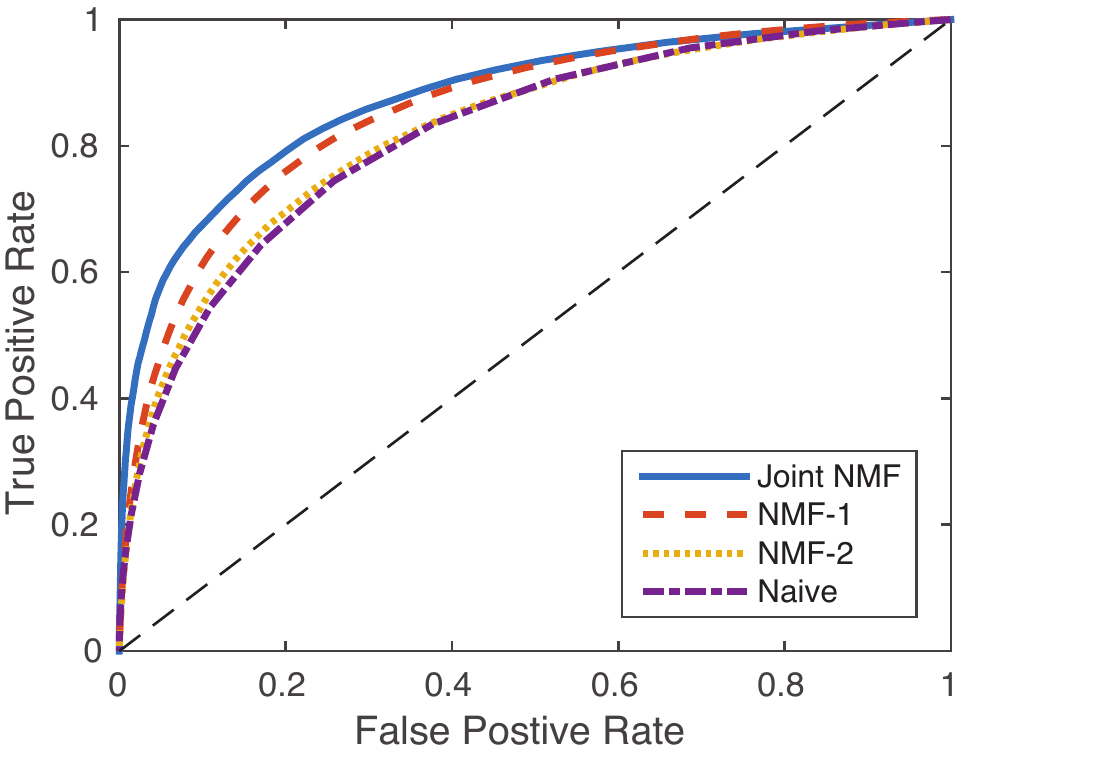}
  \end{minipage}
  \begin{minipage}{\columnwidth}
  \includegraphics[width=\linewidth]{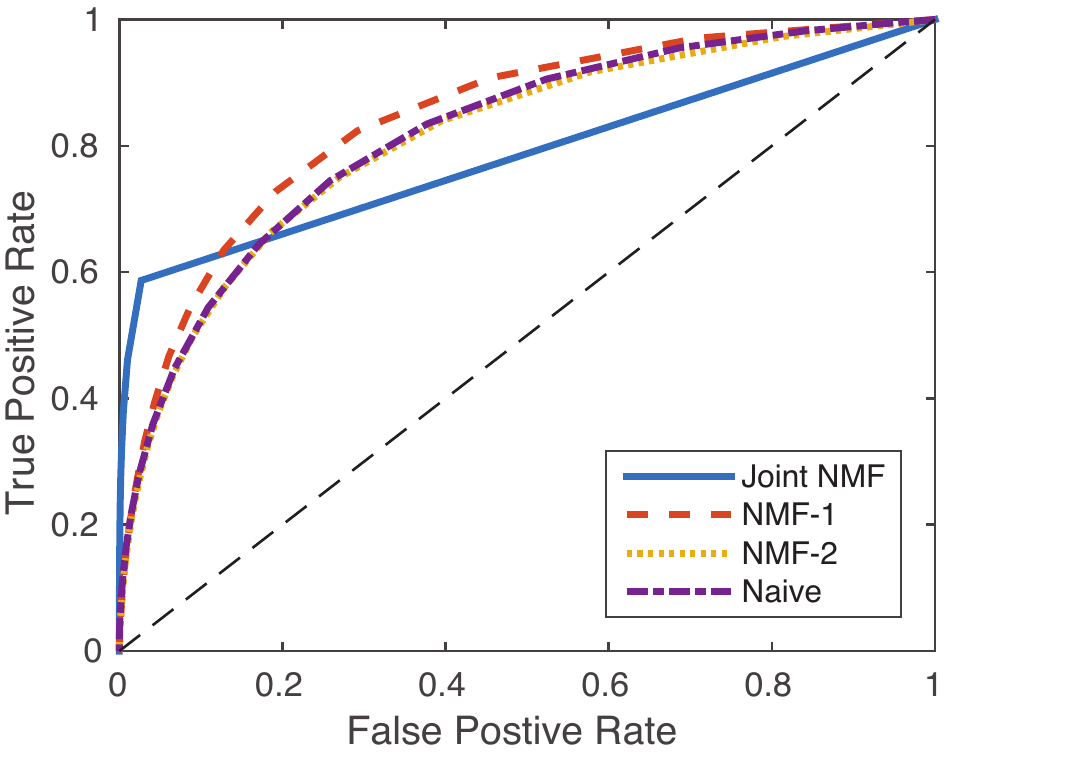}
  \end{minipage}
  \caption{ROC curves for citation recommendation algorithms applied to paper abstract and citation data. The left one uses cosine similarity for the prediction, while the right one uses inner product.}
  \label{fig:cit-abs}
\end{figure*}

We first use paper abstracts as text content. The experiment results are in Figure~\ref{fig:cit-abs}.
We can observe that when cosine similarity is used, joint NMF makes the overall best prediction, and when inner product is used, at certain threshold value joint NMF can achieve relatively high true positive rate with a very low false positive rate. One can choose which one to use based on their requirements.

\begin{figure*}[!tp]
  \centering
  \begin{minipage}{\columnwidth}
  \includegraphics[width=\linewidth]{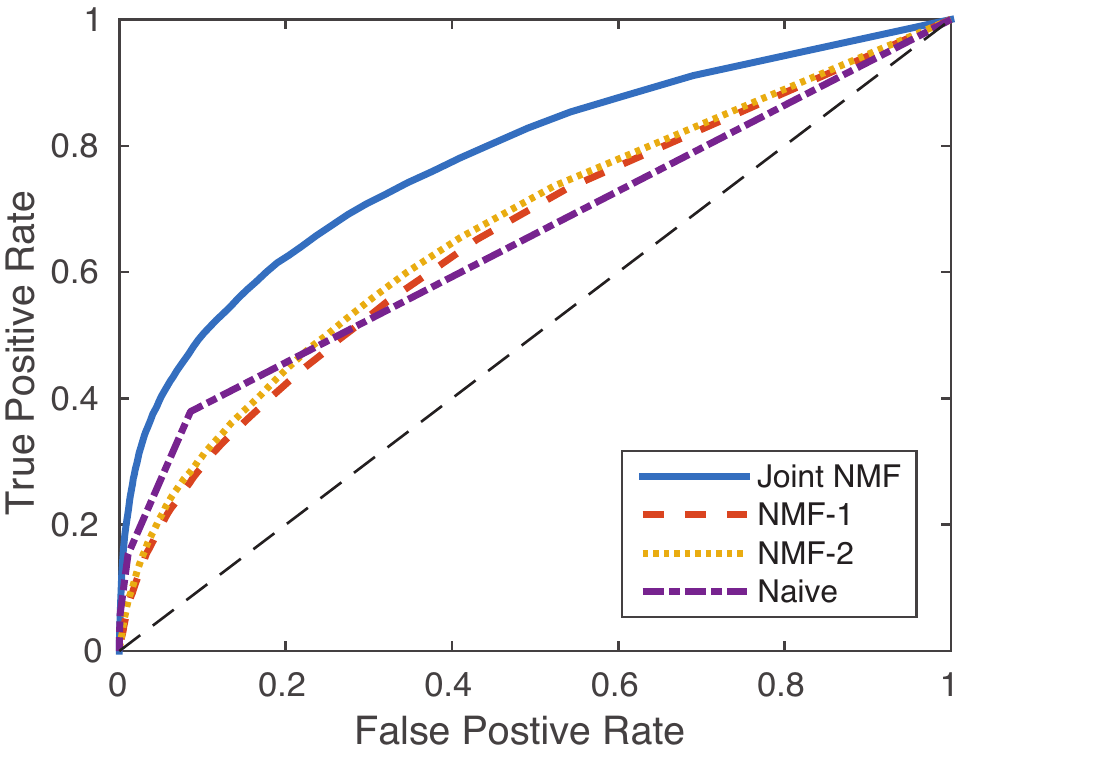}
  \end{minipage}
  \begin{minipage}{\columnwidth}
  \includegraphics[width=\linewidth]{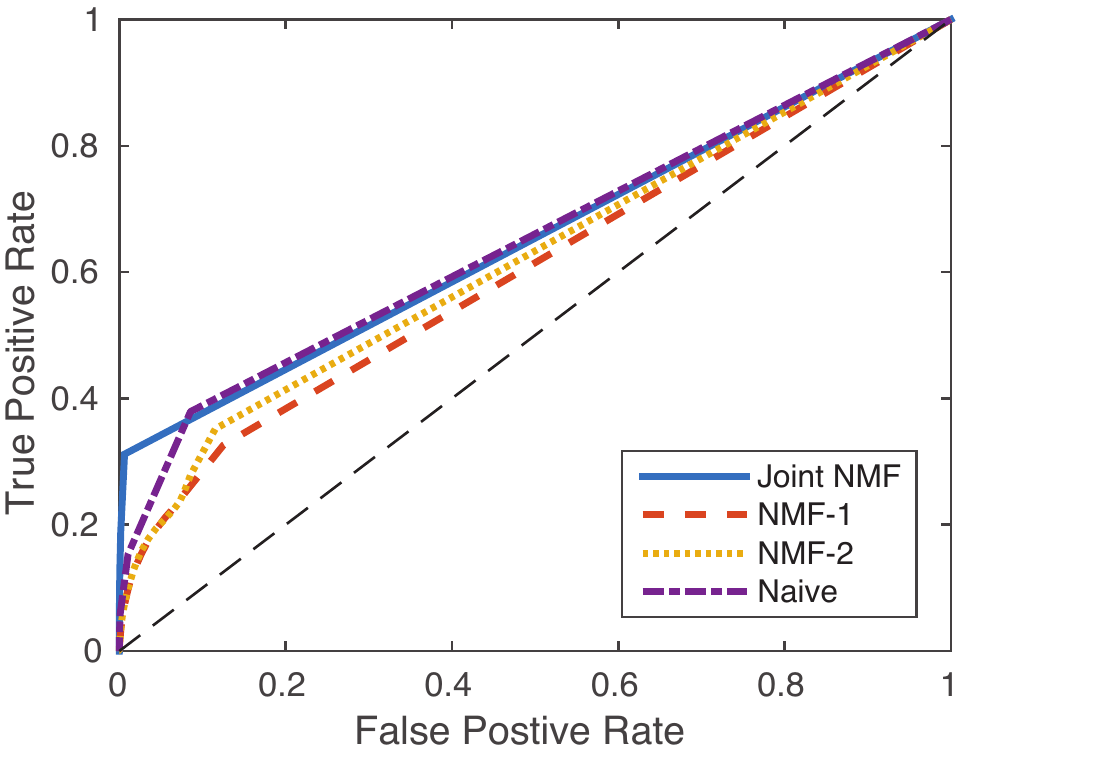}
  \end{minipage}
  \caption{ROC curves for citation recommendation algorithms applied to paper title and citation data. The left one uses cosine similarity for the prediction, while the right one uses inner product.}
  \label{fig:cit-tit}
\end{figure*}

We repeated the experiments using only paper titles as text contents. And similar results are observed, as in Figure~\ref{fig:cit-tit}.
We can observe that even with very little text information (such as paper titles), our method still works.

\subsection{Activity and Leader Detection on Enron email data}
In an organization where different groups of people work on different subjects/have different activities, joint NMF can be used to detect such group structure, reveal the working subject/activities going on and find administrators/leaders in the organization. We assume that (1) within-group communications (e.g. emails) can reflect the subject the team is working on/activities going on and (2) people involved in multiple groups may hold a higher position in the organization, since they may be in charge of these groups. Each communication can be seen as a hypergraph edge that connects all the people involved in the communication and the communication content is the text associated with the edge. Clustering the text data can distinguish and identify different working subjects/activities and clustering the graph data can divide people into workgroups. Joint NMF utilizes both types of data simultaneously and therefore can distinguish different groups of people working on the same subject and different subjects worked on by the same group of people. After the clustering is done, one can count and compare the number of groups/clusters each person belongs to---the more groups a person belongs to, the more likely the person is in a leader or an administrative position.

We use a subset of Enron email data extracted by a group from UC Berkeley \footnote{\url{http://bailando.sims.berkeley.edu/enron_email.html}}, containing 1702 emails. We first construct the term document matrix from email content and the hypergraph incidence matrix from email-sender/recipient relations. The hypergraph has Enron employees as vertices and their emails as edges, and a vertex is connected by an edge if and only if the corresponding employee is the sender or a recipient of the corresponding email. After that, we clean the data by removing terms that appear very infrequently and emails that are too short or duplicated, and extracting the largest connected components of the hypergraph. We then apply tf-idf to the term-document matrix, normalize its columns to have unit 2-norm, obtaining the matrix $X$, and compute $S$ using \eqref{eq:S} in which $M$ is the incidence matrix of the dual hypergraph. Finally, we apply joint NMF with $\alpha = \|X\|_F^2/\|S\|_F^2$ and $\beta=\alpha\|S\|_{max}$ to find 20 groups of employee. Note that since the dual hypergraph is used, the resulting clusters are clusters of emails, instead of clusters of employees. To induce clusters of employees, one simply put employees involved in the same cluster of emails into one employee cluster. In this way, we can actually induce overlapping employee clusters from non-overlapping email clusters. We say an employee has $j$ memberships if the employee belongs to $j$ clusters and count the number of memberships for each employee and list the frequency of each number in Table~\ref{tab:enron-ncluster}.
\begin{table}
  \caption{Frequency of number of memberships}
  \label{tab:enron-ncluster}
  \centering
  \begin{tabular}{rrrrrrrrr}
    \toprule
    \#memberships & 1 &2&3&4&5&6&7&11\\
    \midrule
    \#employees& 1069&149&45&17&8&7&1&1\\
    \bottomrule
  \end{tabular}
\end{table}
We examined the employees that had at least 6 memberships in online news, finding they all held relative high positions in Enron. We list their names and positions in Table~\ref{tab:enron-pos}.
\begin{table}
  \caption{Employees that has $j$ memberships ($j\ge 6$) and their positions in Enron}
  \label{tab:enron-pos}
  \begin{minipage}{\columnwidth}
  \begin{center}
  \setlength{\tabcolsep}{4.5pt}
  \begin{tabular}{rll}
    \toprule
    $j$ & Name & Position in Enron\\
    \midrule
    11 & Steven Kean & Chief of staff\\
    \midrule
    7 & Jeff Dasovich & Governmental affairs executive\\
    \midrule
    &Susan Mara &	California director of Regulatory Affairs\\
    &Richard Shapiro &	VP of regulatory affairs\\
    &Paul Kaufman	&VP of Government Affairs\\
    6&James Steffes	&VP of Government Affairs\\
    &Tim Belden	&Head of trading\\
    &Richard Sanders	&VP  of Enron Whole Sale Services\\
    &Joe Hartsoe	&VP of Federal Regulatory Affairs\\
    \bottomrule
  \end{tabular}
\end{center}
\smallskip
\footnotesize VP: vice president
\end{minipage}
\end{table}
To see the effect of our algorithm on topic modeling, we list some topic keywords for each cluster in Table~\ref{tab:enron-topic}.
\begin{table}
  \caption{Topic keywords of clusters}
  \label{tab:enron-topic}
  \setlength{\tabcolsep}{3pt}
  \begin{tabular}{rl}
    \toprule
    \#	&	Keywords	\\
    \midrule
  0	&	ubs, warburg, forecast, confidential, win	\\
  1	&	blackberry, handheld, wireless	\\
  2	&	california, power, confidential, tariff, pursuant 	\\
  3	&	caiso, refund, ferc, proceedings	\\
  4	&	burrito, peace, things, price, market, board, california	\\
  5	&	document, fax, tonight, sign, back, attach, thanks	\\
  6	&	wholesale, policy, compliance, receipt, legal, service	\\
  7	&	enron, please, know, attach, meeting, contact, call, any, time	\\
  8	&	london, conference, meeting, next, week	\\
  9	&	handheld, blackberry, wireless, agreement, confidential	\\
  10	&	testify, witness, fault, burden, cut, budget	\\
  11	&	california, electricity, energy, price, market, power, rate, bill	\\
  12	&	recommendation, template, participant, management	\\
  13	&	passcode, please, effective, confidential, change	\\
  14	&	stanford, university, expert, try, best, mail, california	\\
  15	&	account, invoice, trust, fund, transfer	\\
  16	&	expense, report, employee, name , approve, amount	\\
  17	&	folder, info, audit, access, apollo, email, sensitivity, server	\\
  18	&	sent, talk, presentation, thanks, infrastructure, amendment	\\
  19	&	hpl, aep, agreement, compete, deal, arrangement 	\\
  \bottomrule
  \end{tabular}
\end{table}
We can observe that some emails are communications about/with other companies and regulatory agencies (0,3,19); some are about administrative tasks or daily work (5,7,8,13,15,16,18); some are about legal issues (6,10); and some are related to the California energy crisis (2,11).

% !TEX root=hybrid.tex
\section{Conclusions and Discussions}
\label{sec:conclusion}
With a simple CLRA formulation in \eqref{eq:jointnmf}, joint NMF is able to solve a variety of problems. The basic application of joint NMF is to cluster hybrid data with both content and connection structure, where the connection structure can be either a graph or a hypergraph, and the content can be associated with either the hypergraph nodes or the edges. When $X$ is any nonnegative feature-data matrix and $S$ is a nonnegative data-data similarity matrix, the joint NMF formulation \eqref{eq:jointnmf} naturally applies without any modification. When there are multiple feature-data matrices $X_1,\ldots,X_p$ and multiple similarity matrices $S_1,\ldots,S_q$, one can extend \eqref{eq:jointnmf} to
\begin{equation*}
  \min_{W_i\geq 0, H\geq 0} \sum_{i=1}^p\alpha_i|| X_i-W_iH ||_F^2 + \sum_{j=1}^q\gamma_j || S_j-H^T H ||_F^2
\end{equation*}
Joint NMF can also be applied to predict paper/patent citations and detect activities and leaders in an organization.

As a hybrid clustering method, joint NMF, with easy-to-set parameters, successfully improves the cluster quality over content-only and connection-only clustering algorithms. It also outperforms one of the leading hybrid clustering methods in the sense of average F1 score. For the performance of pairwise connection prediction, the advantage of joint NMF is its low false positive rate.

In our experiments, joint NMF also shows very good potential for predicting paper/patent citations and activities and leaders in an organization.

Although the current default parameters
($\alpha=\|X\|_F^2/\|S\|_F^2$ and $\beta=\alpha\|S\|_{max}$) for joint NMF are usually good enough, we noticed in our experiments that these are not optimal. We plan to study this further in future research to better understand these parameter values.

Our next research effort, in addition to that noted above, is to accelerate the joint NMF algorithm using a divide-and-conquer approach, as in \cite{kuang_fast_2013-1}. The application of joint NMF to citation recommendation and activity/leader detection will also be further explored and we will conduct more experiments on additional data sets and compare joint NMF with other algorithms in these two areas.

\begin{acks}
This work was supported in part
by the \grantsponsor{SP1796}{National Science Foundation (NSF)}{https://www.nsf.gov/}
grant \grantnum{SP1796}{IIS-1348152} and \grantsponsor{SP242}{Defense Advanced Research Projects Agency (DARPA)}{http://www.darpa.mil/program/xdata} XDATA program grant \grantnum{SP242}{FA8750-12-2-0309}. Any opinions, findings and conclusions or recommendations expressed in this material are those of the authors and do
not necessarily reflect the views of the
NSF or DARPA.
\end{acks}

\bibliographystyle{ACM-Reference-Format}
\bibliography{NMF,hybrid,manual}


\begin{thebibliography}{00}

%%% the Web address.

\ifx \showCODEN    \undefined \def \showCODEN     #1{\unskip}     \fi
\ifx \showDOI      \undefined \def \showDOI       #1{{\tt DOI:}\penalty0{#1}\ }
  \fi
\ifx \showISBNx    \undefined \def \showISBNx     #1{\unskip}     \fi
\ifx \showISBNxiii \undefined \def \showISBNxiii  #1{\unskip}     \fi
\ifx \showISSN     \undefined \def \showISSN      #1{\unskip}     \fi
\ifx \showLCCN     \undefined \def \showLCCN      #1{\unskip}     \fi
\ifx \shownote     \undefined \def \shownote      #1{#1}          \fi
\ifx \showarticletitle \undefined \def \showarticletitle #1{#1}   \fi
\ifx \showURL      \undefined \def \showURL       #1{#1}          \fi
% invisible to TeX
\providecommand\bibfield[2]{#2}
\providecommand\bibinfo[2]{#2}
\providecommand\natexlab[1]{#1}
\providecommand\showeprint[2][]{arXiv:#2}

\bibitem[\protect\citeauthoryear{Bertsekas}{Bertsekas}{1999}]%
        {Bertsekas1999}
\bibfield{author}{\bibinfo{person}{Dimitri Bertsekas}.}
  \bibinfo{year}{1999}\natexlab{}.
\newblock \bibinfo{booktitle}{{\em Nonlinear Programming}}.
\newblock \bibinfo{publisher}{Athena Scientific}.
\newblock


\bibitem[\protect\citeauthoryear{Boyd, Drake, Kuang, and Park}{Boyd
  et~al\mbox{.}}{2016}]%
        {SmallK}
\bibfield{author}{\bibinfo{person}{Richard Boyd}, \bibinfo{person}{Barry
  Drake}, \bibinfo{person}{Da Kuang}, {and} \bibinfo{person}{Haesun Park}.}
  \bibinfo{year}{2016}\natexlab{}.
\newblock \bibinfo{howpublished}{\url{http://smallk.github.io/}}.
  (\bibinfo{date}{June} \bibinfo{year}{2016}).
\newblock


\bibitem[\protect\citeauthoryear{Chang and Blei}{Chang and Blei}{2010}]%
        {chang_hierarchical_2010}
\bibfield{author}{\bibinfo{person}{Jonathan Chang} {and}
  \bibinfo{person}{David~M. Blei}.} \bibinfo{year}{2010}\natexlab{}.
\newblock \showarticletitle{{{HIERARCHICAL RELATIONAL MODELS FOR DOCUMENT
  NETWORKS}}}.
\newblock \bibinfo{journal}{{\em The Annals of Applied Statistics\/}}
  \bibinfo{volume}{4}, \bibinfo{number}{1} (\bibinfo{date}{March}
  \bibinfo{year}{2010}), \bibinfo{pages}{124--150}.
\newblock
\showISSN{1932-6157}


\bibitem[\protect\citeauthoryear{Choo, Lee, Reddy, and Park}{Choo
  et~al\mbox{.}}{2013}]%
        {utopian2013}
\bibfield{author}{\bibinfo{person}{Jaegul Choo}, \bibinfo{person}{Changhyun
  Lee}, \bibinfo{person}{Chandan~K. Reddy}, {and} \bibinfo{person}{Haesun
  Park}.} \bibinfo{year}{2013}\natexlab{}.
\newblock \showarticletitle{UTOPIAN: User-Driven Topic Modeling Based on
  Interactive Nonnegative Matrix Factorization}.
\newblock \bibinfo{journal}{{\em IEEE Transactions on Visualization and
  Computer Graphics\/}} \bibinfo{volume}{19}, \bibinfo{number}{12}
  (\bibinfo{date}{Dec.} \bibinfo{year}{2013}), \bibinfo{pages}{1992--2001}.
\newblock
\showISSN{1077-2626}
\showDOI{%
\url{http://dx.doi.org/10.1109/TVCG.2013.212}}


\bibitem[\protect\citeauthoryear{Cohn and Hofmann}{Cohn and Hofmann}{2001}]%
        {cohn_missing_2001}
\bibfield{author}{\bibinfo{person}{David~A. Cohn} {and} \bibinfo{person}{Thomas
  Hofmann}.} \bibinfo{year}{2001}\natexlab{}.
\newblock \showarticletitle{The {{Missing Link}} - {{A Probabilistic Model}} of
  {{Document Content}} and {{Hypertext Connectivity}}}. In
  \bibinfo{booktitle}{{\em Advances in {{Neural Information Processing
  Systems}} 13}}, \bibfield{editor}{\bibinfo{person}{T.~K. Leen},
  \bibinfo{person}{T.~G. Dietterich}, {and} \bibinfo{person}{V.~Tresp}} (Eds.).
  \bibinfo{publisher}{{MIT Press}}, \bibinfo{pages}{430--436}.
\newblock


\bibitem[\protect\citeauthoryear{Cruz, Bothorel, and Poulet}{Cruz
  et~al\mbox{.}}{2011}]%
        {cruz_entropy_2011}
\bibfield{author}{\bibinfo{person}{J.D. Cruz}, \bibinfo{person}{C. Bothorel},
  {and} \bibinfo{person}{F. Poulet}.} \bibinfo{year}{2011}\natexlab{}.
\newblock \showarticletitle{Entropy Based Community Detection in Augmented
  Social Networks}. In \bibinfo{booktitle}{{\em 2011 {{International
  Conference}} on {{Computational Aspects}} of {{Social Networks}}
  ({{CASoN}})}}. \bibinfo{pages}{163--168}.
\newblock
\showDOI{%
\url{http://dx.doi.org/10.1109/CASON.2011.6085937}}


\bibitem[\protect\citeauthoryear{Drake, Kim, Mallick, and Park}{Drake
  et~al\mbox{.}}{2010}]%
        {DKMP10}
\bibfield{author}{\bibinfo{person}{Barry Drake}, \bibinfo{person}{Jingu Kim},
  \bibinfo{person}{Mahendra Mallick}, {and} \bibinfo{person}{Haesun Park}.}
  \bibinfo{year}{2010}\natexlab{}.
\newblock \showarticletitle{Supervised {R}aman Spectra Estimation based on
  Nonnegative Rank Deficient Least Squares}. In \bibinfo{booktitle}{{\em
  Proceedings 13th International Conference on Information Fusion, Edinburgh,
  UK}}.
\newblock


\bibitem[\protect\citeauthoryear{Elhadi and Agam}{Elhadi and Agam}{2013}]%
        {elhadi_structure_2013}
\bibfield{author}{\bibinfo{person}{Haithum Elhadi} {and} \bibinfo{person}{Gady
  Agam}.} \bibinfo{year}{2013}\natexlab{}.
\newblock \showarticletitle{Structure and {{Attributes Community Detection}}:
  {{Comparative Analysis}} of {{Composite}}, {{Ensemble}} and {{Selection
  Methods}}}. In \bibinfo{booktitle}{{\em Proceedings of the 7th {{Workshop}}
  on {{Social Network Mining}} and {{Analysis}}}} {\em (\bibinfo{series}{SNAKDD
  '13})}. \bibinfo{publisher}{{ACM}}, \bibinfo{address}{New York, NY, USA},
  \bibinfo{pages}{10:1--10:7}.
\newblock
\showISBNx{978-1-4503-2330-7}
\showDOI{%
\url{http://dx.doi.org/10.1145/2501025.2501034}}


\bibitem[\protect\citeauthoryear{Erosheva, Fienberg, and Lafferty}{Erosheva
  et~al\mbox{.}}{2004}]%
        {erosheva_mixed-membership_2004}
\bibfield{author}{\bibinfo{person}{Elena Erosheva}, \bibinfo{person}{Stephen
  Fienberg}, {and} \bibinfo{person}{John Lafferty}.}
  \bibinfo{year}{2004}\natexlab{}.
\newblock \showarticletitle{Mixed-Membership Models of Scientific
  Publications}.
\newblock \bibinfo{journal}{{\em Proceedings of the National Academy of
  Sciences\/}} \bibinfo{volume}{101}, \bibinfo{number}{suppl 1}
  (\bibinfo{date}{June} \bibinfo{year}{2004}), \bibinfo{pages}{5220--5227}.
\newblock
\showISSN{0027-8424, 1091-6490}
\showDOI{%
\url{http://dx.doi.org/10.1073/pnas.0307760101}}


\bibitem[\protect\citeauthoryear{Gruber, Rosen-Zvi, and Weiss}{Gruber
  et~al\mbox{.}}{2008}]%
        {Gruber08}
\bibfield{author}{\bibinfo{person}{Amit Gruber}, \bibinfo{person}{Michal
  Rosen-Zvi}, {and} \bibinfo{person}{Yair Weiss}.}
  \bibinfo{year}{2008}\natexlab{}.
\newblock \showarticletitle{Latent {{Topic Models}} for {{Hypertext}}}. In
  \bibinfo{booktitle}{{\em Proceedings of the {{Twenty}}-{{Fourth Conference
  Annual Conference}} on {{Uncertainty}} in {{Artificial Intelligence}}
  ({{UAI}}-08)}}. \bibinfo{publisher}{{AUAI Press}},
  \bibinfo{address}{Corvallis, Oregon}, \bibinfo{pages}{230--239}.
\newblock


\bibitem[\protect\citeauthoryear{Jin, Gabrys, and Dang}{Jin
  et~al\mbox{.}}{2015}]%
        {jin_combined_2015}
\bibfield{author}{\bibinfo{person}{Di Jin}, \bibinfo{person}{Bogdan Gabrys},
  {and} \bibinfo{person}{Jianwu Dang}.} \bibinfo{year}{2015}\natexlab{}.
\newblock \showarticletitle{Combined Node and Link Partitions Method for
  Finding Overlapping Communities in Complex Networks}.
\newblock \bibinfo{journal}{{\em Scientific Reports\/}}  \bibinfo{volume}{5}
  (\bibinfo{date}{Feb.} \bibinfo{year}{2015}).
\newblock
\showDOI{%
\url{http://dx.doi.org/10.1038/srep08600}}


\bibitem[\protect\citeauthoryear{Kim, He, and Park}{Kim et~al\mbox{.}}{2013}]%
        {kim_algorithms_2013}
\bibfield{author}{\bibinfo{person}{Jingu Kim}, \bibinfo{person}{Yunlong He},
  {and} \bibinfo{person}{Haesun Park}.} \bibinfo{year}{2013}\natexlab{}.
\newblock \showarticletitle{Algorithms for Nonnegative Matrix and Tensor
  Factorizations: A Unified View Based on Block Coordinate Descent Framework}.
\newblock \bibinfo{journal}{{\em Journal of Global Optimization\/}}
  (\bibinfo{date}{March} \bibinfo{year}{2013}).
\newblock
\showISSN{0925-5001, 1573-2916}
\showDOI{%
\url{http://dx.doi.org/10.1007/s10898-013-0035-4}}


\bibitem[\protect\citeauthoryear{Kim and Park}{Kim and Park}{2011}]%
        {kim_fast_2011-1}
\bibfield{author}{\bibinfo{person}{Jingu Kim} {and} \bibinfo{person}{Haesun
  Park}.} \bibinfo{year}{2011}\natexlab{}.
\newblock \showarticletitle{Fast Nonnegative Matrix Factorization: {{An}}
  Active-Set-like Method and Comparisons}.
\newblock \bibinfo{journal}{{\em SIAM Journal on Scientific Computing\/}}
  \bibinfo{volume}{33}, \bibinfo{number}{6} (\bibinfo{year}{2011}),
  \bibinfo{pages}{3261--3281}.
\newblock


\bibitem[\protect\citeauthoryear{Kuang, Choo, and Park}{Kuang
  et~al\mbox{.}}{2015}]%
        {kuang_nonnegative_2015-1}
\bibfield{author}{\bibinfo{person}{Da Kuang}, \bibinfo{person}{Jaegul Choo},
  {and} \bibinfo{person}{Haesun Park}.} \bibinfo{year}{2015}\natexlab{}.
\newblock \showarticletitle{Nonnegative {{Matrix Factorization}} for
  {{Interactive Topic Modeling}} and {{Document Clustering}}}.
\newblock In \bibinfo{booktitle}{{\em Partitional {{Clustering Algorithms}}}},
  \bibfield{editor}{\bibinfo{person}{M.~Emre Celebi}} (Ed.).
  \bibinfo{publisher}{{Springer International Publishing}},
  \bibinfo{pages}{215--243}.
\newblock
\showISBNx{978-3-319-09258-4 978-3-319-09259-1}
\showDOI{%
\url{http://dx.doi.org/10.1007/978-3-319-09259-1_7}}


\bibitem[\protect\citeauthoryear{Kuang and Park}{Kuang and Park}{2013}]%
        {kuang_fast_2013-1}
\bibfield{author}{\bibinfo{person}{Da Kuang} {and} \bibinfo{person}{Haesun
  Park}.} \bibinfo{year}{2013}\natexlab{}.
\newblock \showarticletitle{Fast Rank-2 Nonnegative Matrix Factorization for
  Hierarchical Document Clustering}. In \bibinfo{booktitle}{{\em Proceedings of
  the 19th {{ACM SIGKDD}} International Conference on {{Knowledge}} Discovery
  and Data Mining}}. \bibinfo{publisher}{{ACM}}, \bibinfo{pages}{739--747}.
\newblock


\bibitem[\protect\citeauthoryear{Kuang, Park, and Ding}{Kuang
  et~al\mbox{.}}{2012}]%
        {kuang_symmetric_2012-1}
\bibfield{author}{\bibinfo{person}{Da Kuang}, \bibinfo{person}{Haesun Park},
  {and} \bibinfo{person}{Chris~HQ Ding}.} \bibinfo{year}{2012}\natexlab{}.
\newblock \showarticletitle{Symmetric {{Nonnegative Matrix Factorization}} for
  {{Graph Clustering}}.}. In \bibinfo{booktitle}{{\em {{SDM}}}},
  Vol.~\bibinfo{volume}{12}. \bibinfo{publisher}{{SIAM}},
  \bibinfo{pages}{106--117}.
\newblock


\bibitem[\protect\citeauthoryear{Kuang, Yun, and Park}{Kuang
  et~al\mbox{.}}{2014}]%
        {kuang_symnmf:_2014}
\bibfield{author}{\bibinfo{person}{Da Kuang}, \bibinfo{person}{Sangwoon Yun},
  {and} \bibinfo{person}{Haesun Park}.} \bibinfo{year}{2014}\natexlab{}.
\newblock \showarticletitle{{{SymNMF}}: Nonnegative Low-Rank Approximation of a
  Similarity Matrix for Graph Clustering}.
\newblock \bibinfo{journal}{{\em Journal of Global Optimization\/}}
  \bibinfo{volume}{62}, \bibinfo{number}{3} (\bibinfo{date}{Nov.}
  \bibinfo{year}{2014}), \bibinfo{pages}{545--574}.
\newblock
\showISSN{0925-5001, 1573-2916}
\showDOI{%
\url{http://dx.doi.org/10.1007/s10898-014-0247-2}}


\bibitem[\protect\citeauthoryear{Leskovec and Krevl}{Leskovec and
  Krevl}{2014}]%
        {snapnets}
\bibfield{author}{\bibinfo{person}{Jure Leskovec} {and} \bibinfo{person}{Andrej
  Krevl}.} \bibinfo{year}{2014}\natexlab{}.
\newblock \bibinfo{title}{{SNAP Datasets}: {Stanford} Large Network Dataset
  Collection}.
\newblock \bibinfo{howpublished}{\url{http://snap.stanford.edu/data}}.
  (\bibinfo{date}{June} \bibinfo{year}{2014}).
\newblock


\bibitem[\protect\citeauthoryear{Liu, Wang, Gao, and Han}{Liu
  et~al\mbox{.}}{2013}]%
        {liu_multi-view_2013}
\bibfield{author}{\bibinfo{person}{J. Liu}, \bibinfo{person}{C. Wang},
  \bibinfo{person}{J. Gao}, {and} \bibinfo{person}{J. Han}.}
  \bibinfo{year}{2013}\natexlab{}.
\newblock \showarticletitle{Multi-{{View Clustering}} via {{Joint Nonnegative
  Matrix Factorization}}}.
\newblock In \bibinfo{booktitle}{{\em Proceedings of the 2013 {{SIAM
  International Conference}} on {{Data Mining}}}}. \bibinfo{publisher}{{Society
  for Industrial and Applied Mathematics}}, \bibinfo{pages}{252--260}.
\newblock
\showISBNx{978-1-61197-262-7}


\bibitem[\protect\citeauthoryear{Liu, Niculescu-Mizil, and Gryc}{Liu
  et~al\mbox{.}}{2009}]%
        {liu_topic-link_2009}
\bibfield{author}{\bibinfo{person}{Yan Liu}, \bibinfo{person}{Alexandru
  Niculescu-Mizil}, {and} \bibinfo{person}{Wojciech Gryc}.}
  \bibinfo{year}{2009}\natexlab{}.
\newblock \showarticletitle{Topic-Link {{LDA}}: {{Joint Models}} of {{Topic}}
  and {{Author Community}}}. In \bibinfo{booktitle}{{\em Proceedings of the
  26th {{Annual International Conference}} on {{Machine Learning}}}} {\em
  (\bibinfo{series}{ICML '09})}. \bibinfo{publisher}{{ACM}},
  \bibinfo{address}{New York, NY, USA}, \bibinfo{pages}{665--672}.
\newblock
\showISBNx{978-1-60558-516-1}
\showDOI{%
\url{http://dx.doi.org/10.1145/1553374.1553460}}


\bibitem[\protect\citeauthoryear{Manning, Raghavan, and Sch{\"u}tze}{Manning
  et~al\mbox{.}}{2008}]%
        {manning_introduction_2008}
\bibfield{author}{\bibinfo{person}{Christopher~D. Manning},
  \bibinfo{person}{Prabhakar Raghavan}, {and} \bibinfo{person}{Hinrich
  Sch{\"u}tze}.} \bibinfo{year}{2008}\natexlab{}.
\newblock \bibinfo{booktitle}{{\em Introduction to {{Information Retrieval}}}}.
\newblock \bibinfo{publisher}{{Cambridge University Press}},
  \bibinfo{address}{New York, NY, USA}.
\newblock
\showISBNx{978-0-521-86571-5}


\bibitem[\protect\citeauthoryear{Mei, Cai, Zhang, and Zhai}{Mei
  et~al\mbox{.}}{2008}]%
        {mei_topic_2008}
\bibfield{author}{\bibinfo{person}{Qiaozhu Mei}, \bibinfo{person}{Deng Cai},
  \bibinfo{person}{Duo Zhang}, {and} \bibinfo{person}{ChengXiang Zhai}.}
  \bibinfo{year}{2008}\natexlab{}.
\newblock \showarticletitle{Topic {{Modeling}} with {{Network
  Regularization}}}. In \bibinfo{booktitle}{{\em Proceedings of the 17th
  {{International Conference}} on {{World Wide Web}}}} {\em
  (\bibinfo{series}{WWW '08})}. \bibinfo{publisher}{{ACM}},
  \bibinfo{address}{New York, NY, USA}, \bibinfo{pages}{101--110}.
\newblock
\showISBNx{978-1-60558-085-2}
\showDOI{%
\url{http://dx.doi.org/10.1145/1367497.1367512}}


\bibitem[\protect\citeauthoryear{Nallapati, Ahmed, Xing, and Cohen}{Nallapati
  et~al\mbox{.}}{2008}]%
        {nallapati_joint_2008}
\bibfield{author}{\bibinfo{person}{Ramesh~M. Nallapati}, \bibinfo{person}{Amr
  Ahmed}, \bibinfo{person}{Eric~P. Xing}, {and} \bibinfo{person}{William~W.
  Cohen}.} \bibinfo{year}{2008}\natexlab{}.
\newblock \showarticletitle{Joint {{Latent Topic Models}} for {{Text}} and
  {{Citations}}}. In \bibinfo{booktitle}{{\em Proceedings of the 14th {{ACM
  SIGKDD International Conference}} on {{Knowledge Discovery}} and {{Data
  Mining}}}} {\em (\bibinfo{series}{KDD '08})}. \bibinfo{publisher}{{ACM}},
  \bibinfo{address}{New York, NY, USA}, \bibinfo{pages}{542--550}.
\newblock
\showISBNx{978-1-60558-193-4}
\showDOI{%
\url{http://dx.doi.org/10.1145/1401890.1401957}}


\bibitem[\protect\citeauthoryear{Ruan, Fuhry, and Parthasarathy}{Ruan
  et~al\mbox{.}}{2013}]%
        {ruan_efficient_2013}
\bibfield{author}{\bibinfo{person}{Yiye Ruan}, \bibinfo{person}{David Fuhry},
  {and} \bibinfo{person}{Srinivasan Parthasarathy}.}
  \bibinfo{year}{2013}\natexlab{}.
\newblock \showarticletitle{Efficient {{Community Detection}} in {{Large
  Networks Using Content}} and {{Links}}}. In \bibinfo{booktitle}{{\em
  Proceedings of the {{22Nd International Conference}} on {{World Wide Web}}}}
  {\em (\bibinfo{series}{WWW '13})}. \bibinfo{publisher}{{International World
  Wide Web Conferences Steering Committee}}, \bibinfo{address}{Republic and
  Canton of Geneva, Switzerland}, \bibinfo{pages}{1089--1098}.
\newblock
\showISBNx{978-1-4503-2035-1}


\bibitem[\protect\citeauthoryear{Strehl and Ghosh}{Strehl and Ghosh}{2003}]%
        {strehl_cluster_2003}
\bibfield{author}{\bibinfo{person}{Alexander Strehl} {and}
  \bibinfo{person}{Joydeep Ghosh}.} \bibinfo{year}{2003}\natexlab{}.
\newblock \showarticletitle{Cluster {{Ensembles}} \textemdash{} a {{Knowledge
  Reuse Framework}} for {{Combining Multiple Partitions}}}.
\newblock \bibinfo{journal}{{\em J. Mach. Learn. Res.\/}}  \bibinfo{volume}{3}
  (\bibinfo{date}{March} \bibinfo{year}{2003}), \bibinfo{pages}{583--617}.
\newblock
\showISSN{1532-4435}
\showDOI{%
\url{http://dx.doi.org/10.1162/153244303321897735}}


\bibitem[\protect\citeauthoryear{Sun, Aggarwal, and Han}{Sun
  et~al\mbox{.}}{2012}]%
        {sun_relation_2012}
\bibfield{author}{\bibinfo{person}{Yizhou Sun}, \bibinfo{person}{Charu~C.
  Aggarwal}, {and} \bibinfo{person}{Jiawei Han}.}
  \bibinfo{year}{2012}\natexlab{}.
\newblock \showarticletitle{Relation {{Strength}}-Aware {{Clustering}} of
  {{Heterogeneous Information Networks}} with {{Incomplete Attributes}}}.
\newblock \bibinfo{journal}{{\em Proc. VLDB Endow.\/}} \bibinfo{volume}{5},
  \bibinfo{number}{5} (\bibinfo{date}{Jan.} \bibinfo{year}{2012}),
  \bibinfo{pages}{394--405}.
\newblock
\showISSN{2150-8097}
\showDOI{%
\url{http://dx.doi.org/10.14778/2140436.2140437}}


\bibitem[\protect\citeauthoryear{Tang, Wang, and Liu}{Tang
  et~al\mbox{.}}{2012}]%
        {tang_integrating_2012}
\bibfield{author}{\bibinfo{person}{Jiliang Tang}, \bibinfo{person}{Xufei Wang},
  {and} \bibinfo{person}{Huan Liu}.} \bibinfo{year}{2012}\natexlab{}.
\newblock \showarticletitle{Integrating {{Social Media Data}} for {{Community
  Detection}}}. In \bibinfo{booktitle}{{\em Proceedings of the 2011
  {{International Conference}} on {{Modeling}} and {{Mining Ubiquitous Social
  Media}}}} {\em (\bibinfo{series}{MSM'11})}.
  \bibinfo{publisher}{{Springer-Verlag}}, \bibinfo{address}{Berlin,
  Heidelberg}, \bibinfo{pages}{1--20}.
\newblock
\showISBNx{978-3-642-33683-6}
\showDOI{%
\url{http://dx.doi.org/10.1007/978-3-642-33684-3_1}}


\bibitem[\protect\citeauthoryear{Xu, Yin, Wen, and Zhang}{Xu
  et~al\mbox{.}}{2012}]%
        {admm2012}
\bibfield{author}{\bibinfo{person}{Yangyang Xu}, \bibinfo{person}{Wotao Yin},
  \bibinfo{person}{Zaiwen Wen}, {and} \bibinfo{person}{Yin Zhang}.}
  \bibinfo{year}{2012}\natexlab{}.
\newblock \showarticletitle{An alternating direction algorithm for matrix
  completion with nonnegative factors}.
\newblock \bibinfo{journal}{{\em Frontiers of Mathematics in China\/}}
  \bibinfo{volume}{7}, \bibinfo{number}{2} (\bibinfo{year}{2012}),
  \bibinfo{pages}{365--384}.
\newblock
\showISSN{1673-3576}
\showDOI{%
\url{http://dx.doi.org/10.1007/s11464-012-0194-5}}


\bibitem[\protect\citeauthoryear{Yang and Leskovec}{Yang and Leskovec}{2013}]%
        {yang_overlapping_2013}
\bibfield{author}{\bibinfo{person}{Jaewon Yang} {and} \bibinfo{person}{Jure
  Leskovec}.} \bibinfo{year}{2013}\natexlab{}.
\newblock \showarticletitle{Overlapping Community Detection at Scale: A
  Nonnegative Matrix Factorization Approach}. In \bibinfo{booktitle}{{\em
  Proceedings of the Sixth {{ACM}} International Conference on {{Web}} Search
  and Data Mining}}. \bibinfo{publisher}{{ACM}}, \bibinfo{pages}{587--596}.
\newblock


\bibitem[\protect\citeauthoryear{Yang, Jin, Chi, and Zhu}{Yang
  et~al\mbox{.}}{2009}]%
        {yang_combining_2009}
\bibfield{author}{\bibinfo{person}{Tianbao Yang}, \bibinfo{person}{Rong Jin},
  \bibinfo{person}{Yun Chi}, {and} \bibinfo{person}{Shenghuo Zhu}.}
  \bibinfo{year}{2009}\natexlab{}.
\newblock \showarticletitle{Combining {{Link}} and {{Content}} for {{Community
  Detection}}: {{A Discriminative Approach}}}. In \bibinfo{booktitle}{{\em
  Proceedings of the 15th {{ACM SIGKDD International Conference}} on
  {{Knowledge Discovery}} and {{Data Mining}}}} {\em (\bibinfo{series}{KDD
  '09})}. \bibinfo{publisher}{{ACM}}, \bibinfo{address}{New York, NY, USA},
  \bibinfo{pages}{927--936}.
\newblock
\showISBNx{978-1-60558-495-9}
\showDOI{%
\url{http://dx.doi.org/10.1145/1557019.1557120}}


\bibitem[\protect\citeauthoryear{Zhou, Huang, and Sch{\"o}lkopf}{Zhou
  et~al\mbox{.}}{2007}]%
        {zhou_learning_2007}
\bibfield{author}{\bibinfo{person}{Denny Zhou}, \bibinfo{person}{Jiayuan
  Huang}, {and} \bibinfo{person}{Bernhard Sch{\"o}lkopf}.}
  \bibinfo{year}{2007}\natexlab{}.
\newblock \showarticletitle{Learning with {{Hypergraphs}}: {{Clustering}},
  {{Classification}}, and {{Embedding}}}.
\newblock In \bibinfo{booktitle}{{\em Advances in {{Neural Information
  Processing Systems}} 19}},
  \bibfield{editor}{\bibinfo{person}{B.~Sch{\"o}lkopf}, \bibinfo{person}{J.~C.
  Platt}, {and} \bibinfo{person}{T.~Hoffman}} (Eds.). \bibinfo{publisher}{{MIT
  Press}}, \bibinfo{pages}{1601--1608}.
\newblock


\bibitem[\protect\citeauthoryear{Zhu, Yu, Chi, and Gong}{Zhu
  et~al\mbox{.}}{2007}]%
        {zhu_combining_2007}
\bibfield{author}{\bibinfo{person}{Shenghuo Zhu}, \bibinfo{person}{Kai Yu},
  \bibinfo{person}{Yun Chi}, {and} \bibinfo{person}{Yihong Gong}.}
  \bibinfo{year}{2007}\natexlab{}.
\newblock \showarticletitle{Combining {{Content}} and {{Link}} for
  {{Classification Using Matrix Factorization}}}. In \bibinfo{booktitle}{{\em
  Proceedings of the 30th {{Annual International ACM SIGIR Conference}} on
  {{Research}} and {{Development}} in {{Information Retrieval}}}} {\em
  (\bibinfo{series}{SIGIR '07})}. \bibinfo{publisher}{{ACM}},
  \bibinfo{address}{New York, NY, USA}, \bibinfo{pages}{487--494}.
\newblock
\showISBNx{978-1-59593-597-7}
\showDOI{%
\url{http://dx.doi.org/10.1145/1277741.1277825}}


\end{thebibliography}
\end{document}